\documentclass[letterpaper, 10pt, conference]{ieeeconf}  

\IEEEoverridecommandlockouts                              

\usepackage{graphicx}
\usepackage{mathptmx} 
\usepackage{times} 
\usepackage{amsmath} 
\usepackage{amssymb} 

\usepackage{multirow}
\usepackage{flushend}

\title{\LARGE \bf
A Noise-Robust Turn-Taking System for\\Real-World Dialogue Robots: A Field Experiment
}

\author{Koji Inoue$^{*, 1}$, Yuki Okafuji$^{*, 2, 3}$, Jun Baba$^{2, 3}$, Yoshiki Ohira$^{2, 3}$, Katsuya Hyodo$^{2, 3}$, and Tatsuya Kawahara$^{1}$
\thanks{$^{*}$Koji Inoue and Yuki Okafuji are equal contributors to this work.}
\thanks{This work was supported by JST PRESTO JPMJPR24I4 and JST Moonshot R\&D JPMJPS2011.}
\thanks{$^{1}$Koji Inoue (corresponding author) and Tatsuya Kawahara are with Graduate School of Informatics, Kyoto University, Yoshida-honmachi, Sakyo-ku, Kyoto, Japan, 
        email: {\tt\small inoue.koji.3x@kyoto-u.ac.jp},
        {\tt\small kawahara@i.kyoto-u.ac.jp}}
\thanks{$^{2}$Yuki Okafuji, Jun Baba, Yoshiki Ohira, and Katsuya Hyodo are with CyberAgent, Tokyo, Japan, 
        email: {\tt\small okafuji\_yuki\_xd@cyberagent.co.jp}, \newline
        {\tt\small baba\_jun@cyberagent.co.jp}, \newline
        {\tt\small ohira\_yoshiki@cyberagent.co.jp}, \newline
        {\tt\small hyodo\_katsuya@cyberagent.co.jp}}
\thanks{$^{3}$Yuki Okafuji, Jun Baba, Yoshiki Ohira, and Katsuya Hyodo are with the University of Osaka, Osaka, Japan}
}

\newcommand{\ph}[1]{\phantom{#1}}
\newcommand{\ssp}{\vspace{2mm}}
\newcommand{\eep}{\vspace{-2mm}}
\usepackage{color,soul}

\begin{document}

\maketitle
\thispagestyle{empty}
\pagestyle{empty}

\begin{abstract}
Turn-taking is a crucial aspect of human-robot interaction, directly influencing conversational fluidity and user engagement.
While previous research has explored turn-taking models in controlled environments, their robustness in real-world settings remains underexplored.
In this study, we propose a noise-robust voice activity projection (VAP) model, based on a Transformer architecture, to enhance real-time turn-taking in dialogue robots.
To evaluate the effectiveness of the proposed system, we conducted a field experiment in a shopping mall, comparing the VAP system with a conventional cloud-based speech recognition system.
Our analysis covered both subjective user evaluations and objective behavioral analysis.
The results showed that the proposed system significantly reduced response latency, leading to a more natural conversation where both the robot and users responded faster.
The subjective evaluations suggested that faster responses contribute to a better interaction experience.
\end{abstract}

\renewcommand{\thefootnote}{\fnsymbol{footnote}}
\footnote[0]{This paper has been accepted for presentation at IEEE/RSJ International Conference on Intelligent Robots and Systems 2025 (IROS 2025) and represents the author's version of the work.}
\renewcommand{\thefootnote}{\arabic{footnote}}


\section{Introduction}

In recent years, large language models (LLMs) have advanced rapidly, driving the development of chatbots and spoken dialogue systems capable of facilitating natural human-machine interactions.
By leveraging LLMs, dialogue robots have significantly improved language processing accuracy and response diversity, leading to an increasing number of real-world applications.
In particular, cloud-based speech recognition and response generation services offer high versatility and ease of system updates and extensions.
However, real-world deployment also presents challenges, such as network latency and environmental noise, which cannot always be effectively mitigated.
These factors create significant obstacles to maintaining smooth human-robot interactions in practical settings.

Turn-taking, a fundamental mechanism that determines when a system should start and stop speaking, is one of the most critical functions for ensuring fluid and natural interactions between humans and robots~\cite{skantze2021review}.
However, in real-world environments such as shopping malls and public spaces, where background noise and interruptions are common, conventional speech-recognition-based end-of-utterance detection frequently suffers from misrecognition and delays, leading to turn-taking breakdowns.
As a result, users may interrupt due to impatience, or the robot may respond at inappropriate moments, causing disengagement and diminishing the overall user experience~\cite{tisserand2024unraveling,addlesee2025building}.

This study applies a voice activity projection (VAP) model, based on the Transformer architecture~\cite{erik2022vap}, to enhance turn-taking in dialogue robots.
VAP is a predictive model that estimates near-future voice activity for both conversational participants based on their audio input.
By leveraging these predictions, the robot can regulate its turn-taking behavior more effectively.
The key features of this model include:
(1) End-to-end architecture: Directly predicting turn-taking from audio waveforms, making it robust against speech recognition errors.
(2) Continuous, full-duplex processing: Allowing turn-taking prediction even while the user is still speaking.
(3) Lightweight design: Enabling real-time processing in local environments, even without a GPU~\cite{inoue2024iwsds}.

While prior research has implemented a VAP model in a dialogue robot~\cite{skantze2025hri}, no studies have focused on training a noise-robust VAP model and validating its performance through real-world field experiments.
In this study, we enhance the VAP model by training it on simulated multi-condition data, including noisy environments, to improve turn-taking prediction accuracy in practical settings.

To evaluate the effectiveness of the proposed system, we deployed the robot in a shopping mall and conducted a comparative experiment against a conventional cloud-based speech-to-text system.
The evaluation included subjective user assessments (e.g., smoothness and satisfaction) alongside objective behavioral analyses, such as system and user reaction times.
By examining these factors from multiple perspectives, we aimed to uncover qualitative improvements in interaction quality resulting from the enhanced turn-taking system.
Conducting such a detailed subjective-objective behavioral analysis in real-world field experiments represents a novel approach that has rarely been explored in previous studies.

The contributions of this study are as follows:
\begin{itemize}
    \item We developed a noise-robust VAP model by training it on multi-condition data simulating real-world environments and integrated it into a dialogue robot.
    \item We conducted a real-world field experiment in a public shopping mall, evaluating the robot’s performance through both subjective user assessments and objective, quantitative behavioral analyses.
\end{itemize}

\section{Related work}

Traditional spoken dialogue systems typically determine turn-taking using end-of-utterance detection based on silence or explicit cues, such as pressing a button.  
A common heuristic is to wait for a fixed period of silence (e.g., 500 ms or 1 s) before the system takes its turn~\cite{skantze2021review}.  
While these methods are straightforward and reduce turn-taking complexity, they often result in unnatural interactions.  
Silence-based approaches also struggle in noisy environments, where background noise can interfere with voice activity detection, leading to turn-taking errors.

Recent advancements leverage Transformer-based models to predict turn shifts more accurately~\cite{ekstedt2020turngpt,muromachi2023interspeech,kurata23_interspeech}.
A recent notable innovation is the VAP model, which predicts future speech activity from audio inputs in an end-to-end manner at the frame-wise level~\cite{erik2022vap,erik2022sigdial}.  

In human-robot interaction (HRI), turn-taking is typically based on either silence detection or cloud-based speech recognition systems.  
Several efforts have been made to implement turn-taking models in dialogue robots~\cite{lala2019icmi,yang2022gated}.  
Research indicates that robots equipped with the latest VAP technology can respond more quickly than conventional cloud-based turn-taking systems~\cite{skantze2025hri}.  
However, these systems have primarily been tested in controlled laboratory environments and have not yet demonstrated robust performance in real-world settings, such as shopping malls with ambient noise.  

Within HRI, researchers have extensively studied the effects of turn-taking, or response delay, on various robotic platforms. Most studies focus on subjective evaluations such as user preferences and satisfaction in closed environments such as laboratories~\cite{kanda2007humanoid,shiwa2008quickly,shiomi2017subtle,peng2020understanding,wagner2021address,kim2024impact}. On the other hand, while some studies have observed how delays in robot responses affect user behavior in real-world environments ~\cite{mirnig2015impact,pelikan2023}, research specifically investigating how response speed influences user behavior change in practical settings remains relatively unexplored.

\section{A Noise-robust Turn-taking Model}

This section describes a noise-robust turn-taking model implemented in our robot system used in this study.
We first provide an overview of the VAP model, followed by a newly trained VAP model designed for robustness in noisy real-world environments.

\subsection{Voice Activity Projection}

\begin{figure}[t]
    \centering
    \ssp
    \includegraphics[width=75mm]{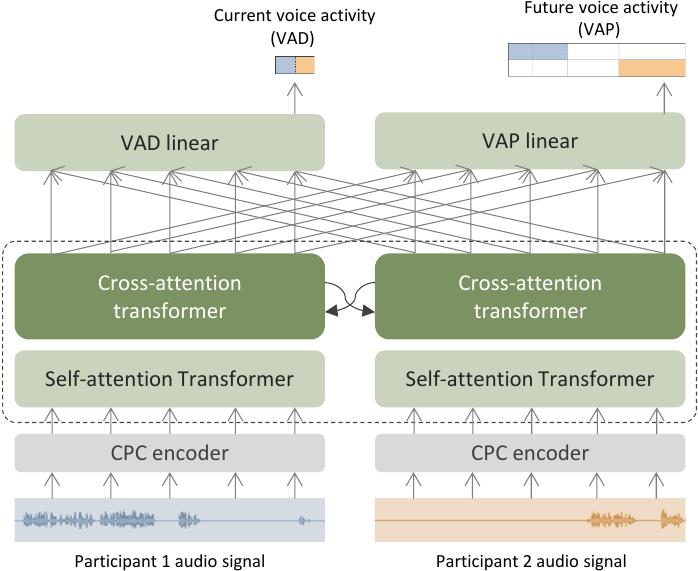}
    \caption{Architecture of VAP}
    \eep
    \label{fig:vap}
\end{figure}

Fig.~\ref{fig:vap} illustrates the architecture of the VAP model. The input consists of separated audio waveforms from two speakers, such as a user and a robot, each encoded using Contrastive Predictive Coding (CPC).
The CPC encoder is pre-trained on 600,000 hours of data from the LibriSpeech dataset~\cite{riviere2020unsupervised}, and its parameters remain frozen during VAP training.
The encoded audio from each speaker is processed by a separate Transformer.
These outputs are then passed through a cross-attention Transformer, allowing the model to attend to interactions between the two speakers.
Finally, a linear layer specific to each task generates the output.

\begin{figure}[t]
    \centering
    \ssp
    \includegraphics[width=70mm]{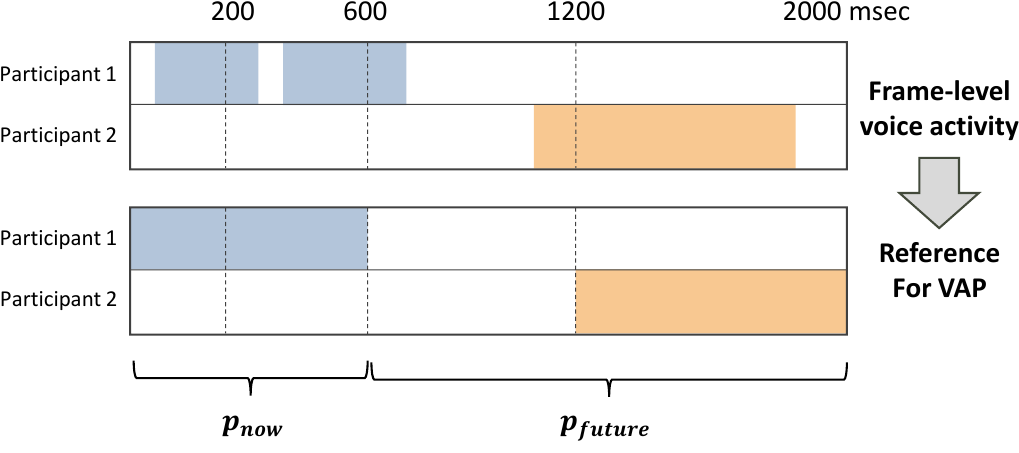}
    \caption{Prediction state of VAP}
    \label{fig:state}
    \eep
\end{figure}

The original VAP model is designed for two tasks: voice activity projection (VAP), which predicts voice activity for the next two seconds, and voice activity detection (VAD), which detects ongoing voice activity.
For the VAP task, the upcoming two seconds are divided into four time bins~\cite{erik2022vap}, as shown in Fig.~\ref{fig:state}.
Since both speakers' voice activity is considered for each bin, there are 256 possible states ($2^{2\times4}$), and the model predicts a 256-class one-hot vector~\cite{inoue2024coling}.
The loss function is defined as follows:
\begin{equation}
L = L_{vap} + L_{vad}, \label{loss}
\end{equation}
where $L_{vap}$ and $L_{vad}$ represent the cross-entropy losses for the VAP and VAD tasks, respectively.

\begin{figure*}[t]
\centering
\ssp
\includegraphics[width=160mm]{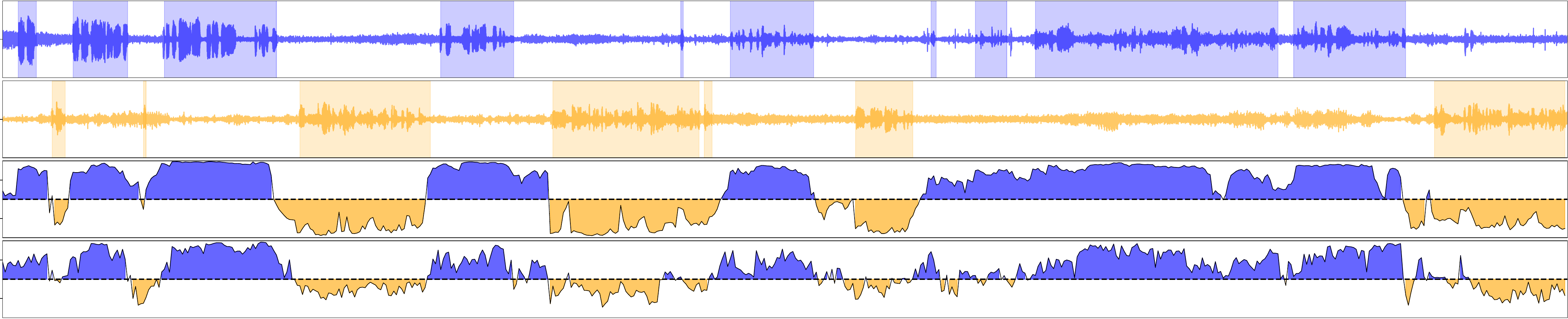}
\caption{Output example of VAP models - Each graph consists of, from top to bottom, input waveforms of both participants with color-highlighted voice activity segments (1st and 2nd), and $p_{now}$ values of multi-condition VAP model (3rd) and conventional clean VAP model (4th)}
\label{fig:vap:output}
\eep
\end{figure*}

In our robotic system, the trained VAP model predicts the end of the user's turn while the user is speaking.
Thus, we input the user's audio into the VAP model and utilize the accumulated output probabilities for short-term VAP predictions within a 0-600 ms window, denoted as $p_{now}$ in the original VAP study~\cite{erik2022vap}.
Note that while the robot’s audio was set to zero in the current setup, future work will incorporate the robot’s text-to-speech (TTS) audio into the model to enable full-duplex interaction.

\begin{table}[t]
    \centering
    \caption{Test loss for VAP task ($L_{vap}$) in different noise levels}
    \begin{tabular}{cccccc}
        \hline
        \multirow{2}{*}{SNR [dB]} & \multicolumn{2}{c}{Travel dataset} & & \multicolumn{2}{c}{ERICA dataset} \\
        \cline{2-3} \cline{5-6}
        & Clean-VAP & MC-VAP & & Clean-VAP  & MC-VAP \\
        \hline
        clean   & 2.37 & 2.48 & & 2.88 & 2.90 \\
        20      & 2.85 & 2.56 & & 3.33 & 3.08 \\
        15      & 2.84 & 2.54 & & 3.55 & 3.14 \\
        10      & 3.02 & 2.55 & & 3.60 & 3.17 \\
        5       & 3.17 & 2.62 & & 4.33 & 3.42 \\
        \hline
    \end{tabular}
    \label{tab:vap:result}
\end{table}

\subsection{Multi-condition Training for VAP}

Previous studies on VAP have primarily trained models on clean speech with minimal background noise, such as that found in telephone or online meeting environments.
However, these models are expected to perform poorly in noisy real-world settings.
To address this limitation, we trained a new VAP model using multi-condition training data to enhance noise robustness.

To simulate multi-condition environments, we superimposed various levels of background noise onto the training data.
The spoken dialogue data used for training included simulated conversations from an online meeting environment designed to mimic travel agency interactions (denoted as {\it Travel})\cite{inaba2023travel}, as well as Wizard-of-Oz dialogue data collected using the android robot ERICA\cite{inoue2016erica} (denoted as {\it ERICA}).
Since a field experiment would be in Japan, those training data were also in the Japanese language.  
In total, these datasets comprised approximately 161 hours of dialogue, which were randomly split into training, validation, and test sets using an 8:1:1 ratio.
The background noise added to these datasets was sourced from CHiME4~\cite{chime4}, DEMAND~\cite{demand}, and MUSAN~\cite{snyder2015musan}, with signal-to-noise ratios (SNRs) of 5, 10, 15, and 20 dB.
During VAP training, both the noise type and SNR level were randomly varied.
To evaluate model performance, we measured the VAP task loss $L_{vap}$ at each SNR level.
Since the VAP model is designed for real-time operation, it generates predictions at a frequency of 10 Hz (i.e., 10 predictions per second) using a 5-second input context sequence for the Transformer model~\cite{inoue2024iwsds}.
As a baseline, we used a model trained without the aforementioned noise augmentation, corresponding to previous studies, denoted as {\it Clean-VAP}.

The results are presented in TABLE~\ref{tab:vap:result}.
In both the {\it Travel} and {\it ERICA} datasets, the baseline model ({\it Clean-VAP}) exhibited an increase in VAP loss as the SNR decreased, indicating degraded performance under noisy conditions.
In contrast, the multi-condition VAP model ({\it MC-VAP}) demonstrated improved robustness, maintaining stable performance even at lower SNR levels.
In this study, we deploy this multi-condition VAP model onto our robot operating in real-world environments and analyze how turn-taking behavior changes during actual interactions.
Fig.~\ref{fig:vap:output} also presents an output example of both VAP models when the SNR was 5 dB.
Overall, the output of the MC model was relatively clearer than that of the clean model, and it was able to predict the next turn-holder earlier.
In some cases where turn-shift occurred between the orange and blue speakers, the clean model struggled to predict the turn shifts clearly, whereas the MC model correctly identified them before they actually happened.
Similar advantages were observed in turn-hold points, where the clean model had difficulty distinguishing between the two speakers, while the MC model accurately predicted that the current speaker would continue holding the turn.
Based on these simulation results, the effectiveness of the MC-VAP model was demonstrated.

\section{Real-field Dialogue Experiment}

We implemented the MC-VAP model in our robot and evaluated its effectiveness through a real-world dialogue experiment.
This section presents the robot and system configuration, followed by the experimental results.

\subsection{Experimental Overview}

To investigate how differences in a robot's speech response speed affect user behavior, we conducted a field experiment comparing two response speed conditions: a faster response condition using the MC-VAP model and a slower response condition using a cloud-based speech recognition system.
While the primary goal of this experiment was to examine the impact of response speed on user behavior, the study also provides insights into the practical advantages of the MC-VAP model in real-world environments.

For this study, we deployed service robots providing route guidance in a shopping mall (LaLaport EXPOCITY\footnote{2-1 Senribanpakukouen, Suita, Osaka, Japan}) and conducted field experiments.
The experiment took place over two days, from February 4 to 5, 2025, with the robots operating between 11 AM and 6 PM.
The baseline method and the proposed method, which will be described later, were each operated for one day.
All visitors were informed about the experiment and the associated video recording through a notification board.
The study was conducted on an opt-out basis, allowing visitors who did not wish to be recorded to decline participation.
This study was approved by the Research Ethics Committee of the University of Osaka (Reference number: R1-5-18).

\subsection{Robot and System}

\begin{figure}[t]
\centering
\ssp
\includegraphics[width=70mm]{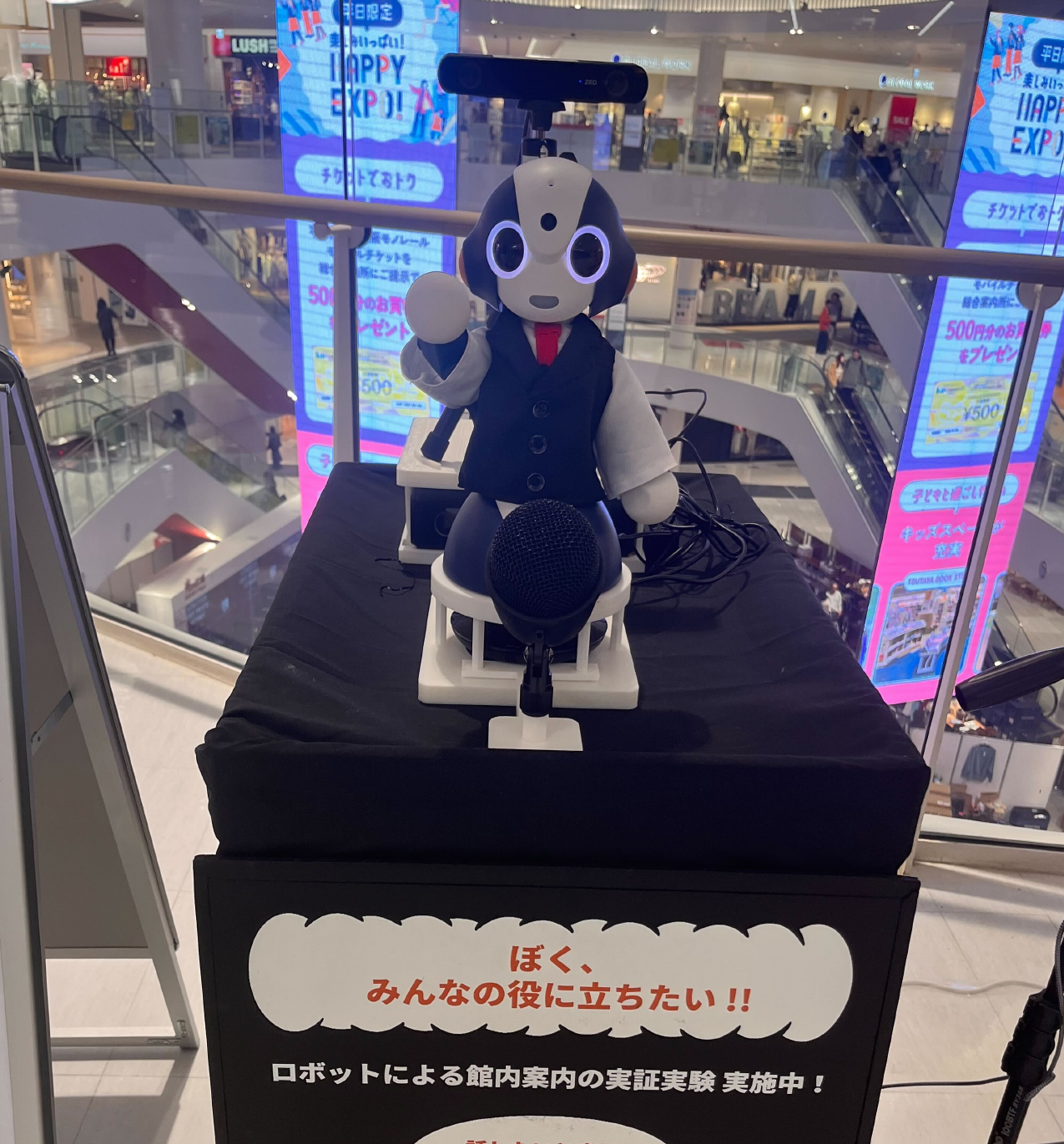}
\caption{The Sota robot, designed for route guidance, equipped with an RGB-D camera}
\label{fig:sota}
\eep
\end{figure}

We developed an interaction system comprising a humanoid social robot and RGB-D image sensors, as shown in Fig.~\ref{fig:sota}.
The social robot used in this experiment was ``Sota'', developed by Vstone Co. Ltd., which has been widely employed in various HRI studies as a service robot (e.g.,~\cite{song2022}).
The humanoid robot stands 0.3 m tall and is equipped with arms that have two degrees of freedom (DOF), a head with three DOF, and body gestures with one DOF.
For 3D imaging, we used the ZED2i sensor, which captures RGB images with a field of view (FOV) of 120 $\times$ 110 $\times$ 70 deg and depth images with an FOV of 81 $\times$ 72 $\times$ 44 deg.
The maximum range of the depth sensor is 35 m.

We developed an autonomous guidance system with four components: Recognition, Turn-taking Detection, Response Generation, and Modality Control.

\subsubsection{Recognition}

The RGB-D camera captures video to detect users' presence and posture in the image~\cite{hyodo2024}.
The system calculates the 3D coordinates of all detected users relative to the robot's coordinate frame.
Speech input is captured via a microphone, processed using the Google Speech-to-Text API (Google STT)~\cite{googlestt}, and then forwarded to the system for further interaction.

\subsubsection{Turn-taking Detection}

After receiving a user’s speech input, the robot employs two distinct systems to determine when the user has completed their utterance.
The first system utilizes the Google STT, which processes speech input and transcribes it into text in real-time.
While streaming, ongoing speech is marked as interim, whereas completed utterances are classified as final.
However, because Google STT relies on a network connection to detect speech completion, network latency directly impacts turn-taking speed.
The second system, the MC-VAP system proposed in this study, predicts speech completion locally. 
Since MC-VAP operates on a local PC, it is expected to facilitate faster turn-taking than Google STT, which depends on cloud-based processing.

In this experiment, we compare a baseline speech response system that relies solely on Google STT with a hybrid system that combines MC-VAP and Google STT.
This comparison aims to examine how response speed influences user behavior.
The proposed hybrid system is employed because MC-VAP alone cannot accurately predict utterance completion for all speech inputs.
In cases where MC-VAP fails, the system defaults to Google STT.

\subsubsection{Response Generation}
The system includes a GPT-4o-mini-powered generative model~\cite{openai2024gpt4omini}.
The generative model was used to generate Sota's responses to user's speech input.
The system incorporated a database containing comprehensive mall information for providing route guidance.
When the system detected the user’s speech completion, the system combined the dialogue history with the mall information to generate an appropriate response.

\subsubsection{Modality Control}
The robot's speech synthesis is generated using VOICEBOX~\cite{voicebox}, with the Japanese character voice ``Zundamon''. When the robot is speaking, the microphone input is turned off because the robot's own speech is also input by the microphone. This means that the system is designed so that the user cannot barge in while the robot is speaking. In addition to responding verbally, Sota also gazed at the closest user detected by the RGB-D camera. This function is intended to increase the conversation opportunity by drawing the user's attention to the robot~\cite{Okafuji2020}.

\subsection{Measurement}

We measured not only subjective evaluation by the users but also objective behavior analysis to investigate how differences in a robot’s response speed affect user behavior.

\paragraph{Subjective Evaluation}

After each interaction, we asked the users to rate 15 items listed in TABLE~\ref{tab:subjective} on a seven-point Likert scale.
These items were designed based on existing metrics, such as User Experience Questionnaire (UEQ)~\cite{laugwitz2008construction} and Godspeed~\cite{bartneck2009measurement}.
Additionally, we created some original items to complement these metrics regarding the system response.

\paragraph{Behavior Analysis}

The recorded dialogue logs and video-recorded interaction data were analyzed retrospectively.  
First, the robot’s response time was measured. This was defined as the time between obtaining the final speech recognition result of the user's utterance and the robot beginning its response.  
Additionally, the user’s response time was also measured.
In this case, it was defined as the time between the end of the robot’s utterance and the first speech recognition result of the user’s subsequent utterance.  
The visitors’ behavior during the conversation was also analyzed.
Specifically, the following were manually counted: interaction time, the number of times the visitor rephrased their utterance, the number of times the robot and the visitor spoke simultaneously (speech collisions), the total number of visitor utterances, the number of questions asked by the visitor, and the number of times the visitor left the conversation midway, listed in TABLE~\ref{tab:behavior}.

\subsection{Result}

Each condition of the proposed and baseline systems could be evaluated by 15 different visitors, respectively.
Table~\ref{tab:subjective} shows the results of the users' subjective evaluation scores.
These results were tested for homogeneity of variance, and for items where homogeneity of variance was found, a Student's $t$-test was conducted, while for items where homogeneity of variance was not found, a Welch t-test was conducted to perform statistical testing.
The results showed that the proposed model had a significantly more positive impact on the three items of Smooth conversation, Ease of use, and Fulfillment of expectations.

\begin{table}[t]
    \centering
    \ssp
    \caption{Average subjective evaluation scores}
    \begin{tabular}{lccl}
        \hline
        \multicolumn{1}{c}{Item} & Baseline & Proposed & $p$-value \\
        \hline
        Smooth conversation      & 4.67 & 6.13 & .007 ** \\
        Appropriate speech content {\dag}              & 5.80 & 6.13 & .442 \\
        Naturalness of utterance {\dag}                & 5.87 & 5.60 & .509 \\
        Intelligent {\dag}                            & 5.67 & 6.00 & .437 \\
        Likability {\dag}                             & 6.40 & 6.60 & .514 \\
        Satisfaction {\dag}                            & 5.73 & 5.93 & .696 \\
        Practicality                            & 5.67 & 5.73 & .895 \\
        Ease of use                             & 5.73 & 6.67 & .048 * \\
        Fulfillment of expectations             & 5.47 & 6.53 & .045 * \\
        Valuable {\dag}                                & 5.87 & 6.13 & .533 \\
        Innovative {\dag}                              & 5.80 & 5.67 & .830 \\
        Able to converse at a good pace {\dag}         & 5.00 & 5.67 & .221 \\
        Frustrated with response speed {\dag}          & 3.07 & 2.60 & .583 \\
        Would like to use again {\dag}                 & 5.27 & 5.73 & .413 \\
        Would recommend to others {\dag} {\ddag}     & 7.13 & 8.13 & .190 \\
        \hline
        \multicolumn{4}{r}{{\dag} Homogeneity of variance, {\ddag} 10-point scale (1-10)} \\
        \multicolumn{4}{r}{*~$p<.05$, **~$p<.01$} \\
    \end{tabular}
    \label{tab:subjective}
    \eep
\end{table}

\begin{figure}[t]
    \ssp
    \centering
    \includegraphics[width=\linewidth]{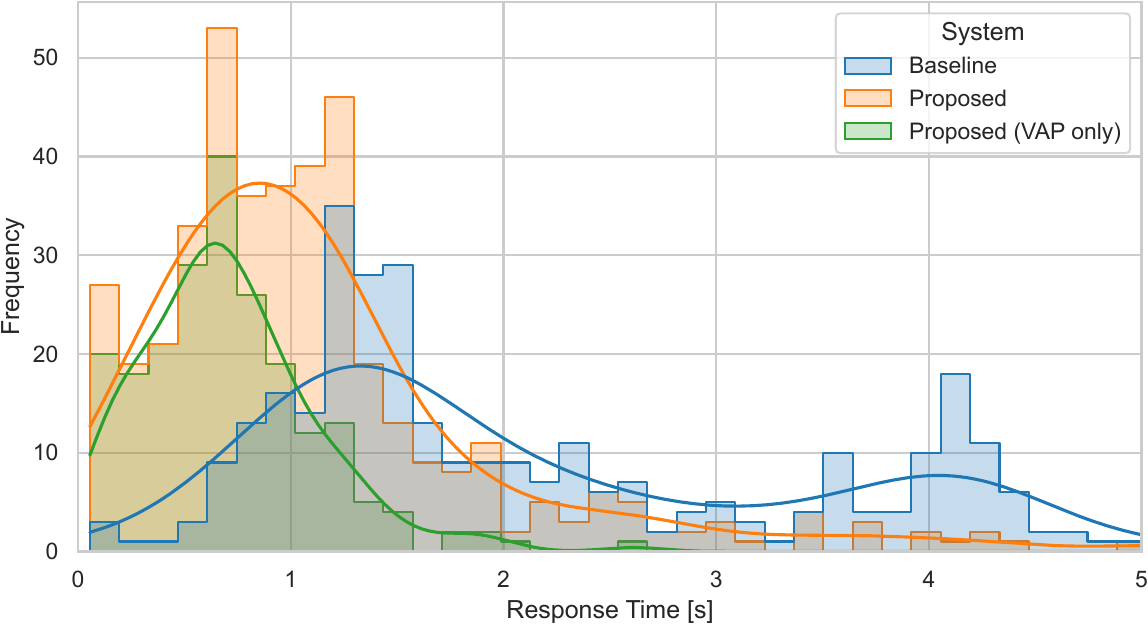}
    \caption{Distribution of robot response time}
    \label{fig:dist:response:robot}
    \eep
\end{figure}

Fig.~\ref{fig:dist:response:robot} illustrates the distribution of the robot’s response time.
Note that these results include interactions not only with the 15 visitors who completed the subjective evaluations but also with additional participants, totaling 71 individuals per condition.
The baseline system, which relies solely on Google STT, exhibits a broad distribution of response times, frequently exceeding 2 seconds due to network latency, with an average response time of 2.14 seconds.
In contrast, the proposed system, which incorporates the MC-VAP model, significantly reduces response latency, with an average response time of 1.15 seconds.
Within this system, turn-taking decisions were determined using MC-VAP in 51.0\% of cases and Google STT in 49.0\%, indicating that the system dynamically switches between these two methods.
To further analyze the impact of MC-VAP, we extracted only the instances where MC-VAP was used and plotted them separately as the ``Proposed (VAP only)'' condition.
This subset of data reveals that when the VAP model alone was responsible for turn-taking decisions, the average response time was further reduced to 0.71 seconds.
A Mann-Whitney U test confirmed that the differences among conditions were all statistically significant ($p<.01$).
These results highlight the effectiveness of MC-VAP in enabling faster and more fluid interactions in real-world dialogue settings.

\begin{figure}[t]
    \ssp
    \centering
    \includegraphics[width=\linewidth]{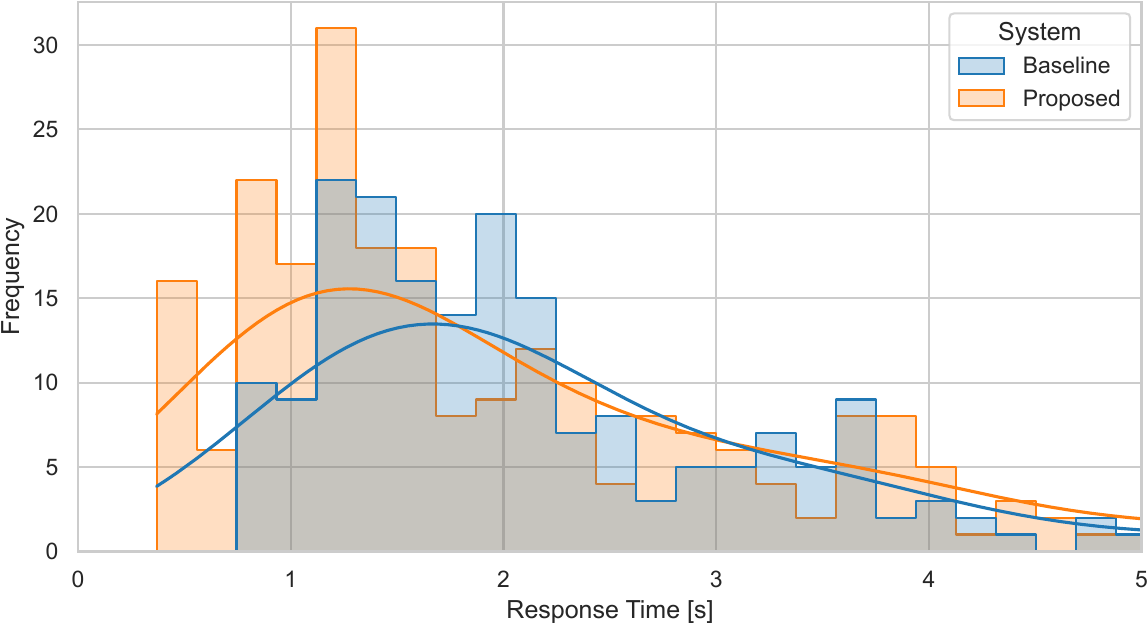}
    \caption{Distribution of user response time}
    \label{fig:dist:response:user}
    \eep
\end{figure}

Fig.~\ref{fig:dist:response:user} presents the distribution of users’ response times.
The baseline system shows a relatively wide distribution of response times, with an average response time of 2.61 seconds.
In contrast, the proposed system results in a slightly faster, with an average response time of 2.35 seconds.
This result suggests that users adapt more quickly to the system’s faster responses, facilitating a more natural and fluid interaction.
A Mann-Whitney U test confirmed that the difference in user response times between the baseline and proposed systems was statistically significant ($p<.01$).
This indicates that reducing the robot’s response time has a measurable impact on user behavior, encouraging more immediate turn-taking and engagement.

\begin{table}[t]
    \centering
    \ssp
    \caption{Average values of user behaviors}
    \begin{tabular}{lccc}
        \hline
        \multicolumn{1}{c}{Item} & Baseline & Proposed & $p$-value \\
        \hline
        Interaction time [s]    & 53.74 & 55.51 & .819 \\
        Number of rephrases              & \ph{0}0.65 & \ph{0}0.55 & .596 \\
        Number of collisions {\dag}      & \ph{0}0.35 & \ph{0}0.34 & .863 \\
        Number of visitor utterances     & \ph{0}5.67 & \ph{0}6.00 & .448 \\
        Number of visitor questions      & \ph{0}1.30 & \ph{0}1.28 & .963 \\
        Visitor left [\%]       & 10.81 & 14.08 & - \\
        \hline
        \multicolumn{4}{r}{{\dag} Homogeneity of variance} 
    \end{tabular}
    \label{tab:behavior}
    \eep
\end{table}

TABLE~\ref{tab:behavior} presents the objective analysis of user behaviors under the baseline and proposed system conditions.
These results, as well as the users' subjective evaluation, were tested for homogeneity of variance using Student's or Welch t-tests.
Overall, the results indicate that while the proposed system improved response speed, it did not lead to statistically significant changes in users' behaviors.

\section{Conclusion}

This study evaluated the effectiveness of the multi-condition voice activity projection (MC-VAP) model through a field experiment in a shopping mall.
The results confirmed that MC-VAP significantly reduced response latency while maintaining stable performance.
The results also showed that users responded more quickly under the proposed system, leading to a more natural conversation.
These improvements were reflected in higher subjective ratings for ``Smoothness of conversation,'' ``Ease of use,'' and ``Fulfillment of expectations,'' suggesting that faster response speeds contribute to a better user experience. 
However, the objective behavioral analysis, except for response time, did not change significantly.
This suggests that while reducing response latency enhances perceived interaction quality, it does not necessarily lead to behavioral changes in real-world settings.

This study was conducted in a single real-world setting, and future research should explore different environments to assess generalizability, including language differences~\cite{inoue2024coling}.
Additionally, while the findings indicate that faster response speeds improve conversational smoothness, further investigation is needed to determine their influence on actual user behaviors, which did not show significant improvement in this experiment.
A deeper understanding of these behavioral changes will help refine real-world turn-taking strategies for dialogue robots, ensuring both subjective and behavioral enhancements in human-robot interactions.

\bibliographystyle{IEEEtran}
\bibliography{IROS2025_KI}

\end{document}